%% file: main.tex
\documentclass{article}

\usepackage[numbers]{natbib}

\usepackage[final]{neurips_2021}

\usepackage[utf8]{inputenc} 
\usepackage[T1]{fontenc}    
\usepackage{url}            
\usepackage{booktabs}       
\usepackage{amsfonts}       
\usepackage{nicefrac}       
\usepackage{microtype}      

\usepackage{times}
\usepackage{graphicx}
\usepackage{amsmath}
\usepackage{amssymb}
\usepackage{booktabs,arydshln}
\usepackage{xspace}
\usepackage{multirow}
\usepackage[dvipsnames]{xcolor}

\usepackage[pagebackref=true,breaklinks=true,letterpaper=true,colorlinks,bookmarks=false,citecolor=ForestGreen]{hyperref}

\makeatletter\renewcommand\paragraph{\@startsection{paragraph}{4}{\z@}
  {.5em \@plus1ex \@minus.2ex}{-.5em}{\normalfont\normalsize\bfseries}}\makeatother

\makeatletter
\def\adl@drawiv#1#2#3{%
        \hskip.5\tabcolsep
        \xleaders#3{#2.5\@tempdimb #1{1}#2.5\@tempdimb}%
                #2\z@ plus1fil minus1fil\relax
        \hskip.5\tabcolsep}
\newcommand{\cdashlinelr}[1]{%
  \noalign{\vskip\aboverulesep
           \global\let\@dashdrawstore\adl@draw
           \global\let\adl@draw\adl@drawiv}
  \cdashline{#1}
  \noalign{\global\let\adl@draw\@dashdrawstore
           \vskip\belowrulesep}}
           
\newcommand{\ours}{VATT\xspace}
\input{math_commands.tex}

\title{VATT: Transformers for Multimodal Self-Supervised Learning from Raw Video, Audio and Text}

\author{Hassan Akbari\thanks{Work done during an internship at Google.} \\
Columbia University \\
\texttt{\tt\small ha2436@columbia.edu} \\
\And
Liangzhe Yuan \\
Google \\
\texttt{\tt\small lzyuan@google.com} \\
\And
Rui Qian\footnotemark[1] \\
Cornell University \\
\texttt{\tt\small rq49@cornell.edu} \\
\And
Wei-Hong Chuang \\
Google \\
\texttt{\tt\small whchuang@google.com} \\
\And
Shih-Fu Chang \\
Columbia University \\
\texttt{\tt\small sc250@columbia.edu} \\
\And
Yin Cui \\
Google \\
\texttt{\tt\small yincui@google.com} \\
\And
Boqing Gong \\
Google \\
\texttt{\tt\small bgong@google.com} \\
}

\begin{document}

\maketitle

\begin{abstract}
  We present a framework for learning multimodal representations from unlabeled data using convolution-free Transformer architectures. Specifically, our \textbf{V}ideo-\textbf{A}udio-\textbf{T}ext \textbf{T}ransformer (\textbf{VATT}) takes raw signals as inputs and extracts multimodal representations that are rich enough to benefit a variety of downstream tasks. We train VATT end-to-end from scratch using multimodal contrastive losses and evaluate its performance by the downstream tasks of video action recognition, audio event classification, image classification, and text-to-video retrieval. Furthermore, we study a modality-agnostic, single-backbone Transformer by sharing weights among the three modalities. 
  We show that the convolution-free VATT outperforms state-of-the-art ConvNet-based architectures in the downstream tasks. 
  Especially, VATT's vision Transformer achieves the top-1 accuracy of 82.1\% on Kinetics-400, 83.6\% on Kinetics-600, 72.7\% on Kinetics-700, and 41.1\% on Moments in Time, new records while avoiding supervised pre-training.
  Transferring to image classification leads to $78.7\%$ top-1 accuracy on ImageNet compared to $64.7\%$ by training the same Transformer from scratch, showing the generalizability of our model despite the domain gap between videos and images.
  VATT's audio Transformer also sets a new record on waveform-based audio event recognition by achieving the mAP of 39.4\% on AudioSet without any supervised pre-training. VATT's source code is publicly available.\footnote{\href{https://github.com/google-research/google-research/tree/master/vatt}{https://github.com/google-research/google-research/tree/master/vatt}}
\end{abstract}

\section{Introduction}

\begin{figure*}[t]
\centering
   \includegraphics[width=0.95\linewidth]{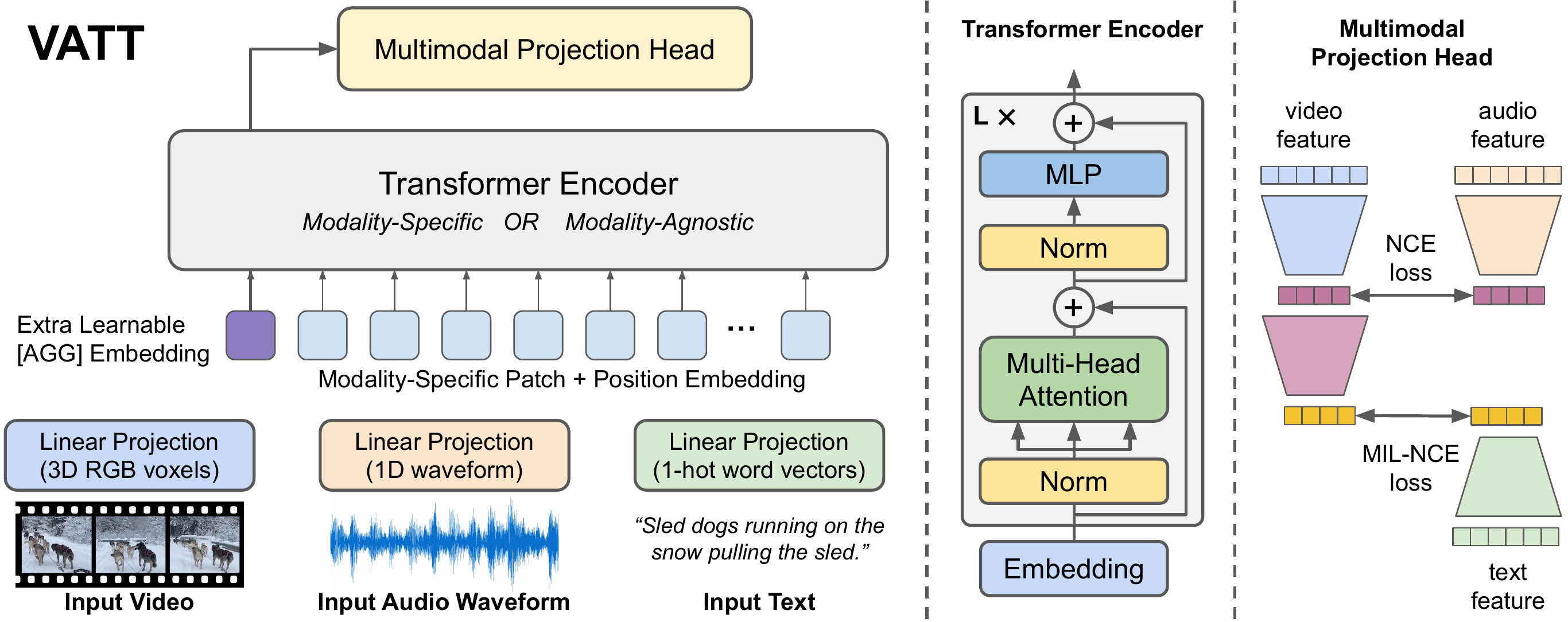}
   \caption{\textbf{Overview of the \ours architecture and the self-supervised, multimodal learning strategy}. VATT linearly projects each modality into a feature vector and feeds it into a Transformer encoder. We define a semantically hierarchical common space to account for the granularity of different modalities and employ the Noise Contrastive Estimation (NCE) to train the model.}
\label{fig:overview}
\vspace{-4mm}
\end{figure*}

Convolutional neural networks (CNNs)~\cite{lecun1998gradient,krizhevsky2012imagenet} have triumphed over various computer vision tasks. The inductive bias induced by convolutions, namely translation invariance and locality, are proven effective for the visual data. In the meantime, however, we witness in the natural language processing (NLP) community a paradigm shift from the models with strong inductive biases, such as recurrent neural networks~\cite{hochreiter1997long,bahdanau2014neural} and CNNs~\cite{zhang2015character,gehring2017convolutional}, to more general architectures constructed upon self-attention. Particularly, Transformers~\cite{vaswani2017attention} have become the de facto model architecture for NLP tasks~\cite{devlin2018bert,radford2018improving,radford2019language,brown2020language}. Pre-training a Transformer on large text corpora followed by fine-tuning gives rise to state-of-the-art results for different downstream tasks. 

In view of the success of the attention mechanism in NLP, there has been a rich line of works exploring its potential in computer vision. Early work studied hybrid models consisting of both convolutions and attention modules~\cite{wang2017residual,woo2018cbam,girdhar2017attentional,zhang2019residual}. Recent studies showed that convolution-free, specially designed all-attention models can match CNNs' performance on image recognition tasks~\cite{zhao2020exploring,hu2019local,ramachandran2019stand}. Most recently, ~\cite{dosovitskiy2021an} achieved impressive performance on several image recognition tasks, including ImageNet~\cite{deng2009imagenet}, using a pre-trained Transformer with minimal architecture changes. Their work delivered a compelling message that ``large scale (supervised) training trumps inductive bias (for image classification).'' This conclusion was further extended to video recognition tasks by~\cite{bertasius2021space,vivit}. 

However, the large-scale supervised training of Transformers is essentially troubling for two main reasons. First, it rules out the much larger other part of ``big visual data,'' i.e, the vast amount of unlabeled, unstructured visual data. As a result, the supervised training strategy could produce biased systems that require even more labeled data to correct their biases. Second, this strategy fundamentally limits the application scope of Transformers in computer vision because it is costly and extremely time-consuming to collect enough labeled images or videos for training the millions of parameters, choosing hyper-parameters, and validating their expected generalization. 

Hence, this work poses another pressing question about the Transformers that take raw signals as input. \emph{How to empower them with large-scale, unlabeled visual data?} To answer this question, we draw insights from NLP. BERT~\cite{devlin2018bert} and GPT~\cite{radford2018improving, radford2019language, brown2020language}
use masked language modeling as their pre-training tasks. Natural languages are organic supervision for Transformers. They sequentially place words, phrases, and sentences into context, granting them semantics and syntax. For visual data, \emph{the most organic supervision is arguably the multimodal videos.}  They are abundantly available in the digital world, and their temporal, cross-modality regulation, and therefore supervision, requires no human annotation. The extreme scale of multimodal videos is potentially capable to teach Transformers necessary priors, as opposed to predefined inductive biases, to model the visual world.

To this end, we study self-supervised, multimodal pre-training of three Transformers~\cite{vaswani2017attention}, which take as input the raw RGB frames of internet videos, audio waveforms, and text transcripts of the speech audio, respectively. We call the video, audio, text Transformers \ours. Figure~\ref{fig:overview} illustrates the architecture. VATT borrows the exact architecture from BERT~\cite{devlin2018bert} and ViT~\cite{dosovitskiy2021an} except the layer of tokenization and linear projection reserved for each modality separately. This design shares the same spirit as ViT that we make the minimal changes to the architecture so that the learned model can transfer its weights to various frameworks and tasks. Furthermore, the self-supervised, multimodal learning strategy resonates the spirit of BERT and GPT that the pre-training requires minimal human curated labels. 

We evaluate the pre-trained Transformers on a variety of downstream tasks: \emph{image classification, video action recognition, audio event
classification, and zero-shot text-to-video retrieval}. Fine-tuning the vision-modality Transformer on ImageNet~\cite{deng2009imagenet} obtains the top-1 accuracy of $78.7\%$, which is comparable to $79.9\%$ achieved by ViT. This result is especially appealing considering the domain gap between videos and images, and that ViT is pre-trained using a large-scale, human-curated image dataset. Furthermore, we set new records on Kinetics-400~\cite{kinetics400}, Kinetics-600~\cite{kinetics600}, Moments in Time~\cite{monfort2019moments}, and AudioSet~\cite{audioset} without supervised pre-training.

Our VATT results, along with others reported for NLP tasks~\cite{devlin2018bert,brown2020language}, image recognition~\cite{dosovitskiy2021an}, semantic segmentation~\cite{zheng2020rethinking}, point cloud classification~\cite{zhao2020point}, and action recoginition~\cite{bertasius2021space}, demonstrate that Transformer is a versatile general-purpose architecture for different types of data. 

To move one step forward, we challenge the Transformers in VATT by a seemingly too strong constraint: sharing weights among the video, audio, and text modalities. The idea is to test whether there exists a single, general-purpose model for all the modalities --- of course, they still have their own layers of tokenization and linear projection. Preliminary results are encouraging. This modality-agnostic Transformer is on par with three modality-specific ones of slightly smaller sizes.

Finally, another contribution of this work is DropToken, a simple and yet effective technique to reduce the training complexity with a minor reduction of the end Transformers' performance. DropToken randomly drops a portion of the video and audio tokens from each input sequence during training, allowing for high-resolution inputs and leveraging their abundance. This is significant for Transformers because their computational complexity is quadratic with respect to the number of input tokens. 

\section{Related work}
\subsection{Transformers in Vision}
Transformer was originally built for NLP tasks~\cite{vaswani2017attention} and the design of multi-head attention shows its effectiveness on modeling long-term correlation of words. A few attempts have been made to use Transformer for vision tasks like image super-resolution~\cite{yang2020learning}, object detection~\cite{carion2020end} and multimodal video understanding~\cite{sun2019learning, chen2020uniter, luo2020univilm}. However these methods still rely on the feature extracted by CNNs.  Recently, \cite{dosovitskiy2021an} proposes a set of convolution-free vision Transformers which directly work on raw images and obtain competitive performance with CNNs. \cite{touvron2020training} improves the training data efficiency of~\cite{dosovitskiy2021an} by using stronger data augmentations and knowledge distillation. Since then, the pure Transformer design has been adopted to various vision tasks including semantic segmentation~\cite{zheng2020rethinking}, point cloud classification~\cite{zhao2020point}, action recoginition~\cite{bertasius2021space,sharir2021image,vivit}. To the best of our knowledge, our \ours is the first Transformer model on raw multimodal inputs of video, audio and text.

\subsection{Self-Supervised Learning}
\paragraph{Single vision modality.} 
Early work of self-supervised visual representation learning usually learns from unlabeled images via manually specified pretext tasks, like auto-encoding~\cite{pathak2016context,zhang2016colorful,zhang2017split}, patch location prediction~\cite{doersch2015unsupervised}, solving jigsaw puzzles~\cite{noroozi2016unsupervised}, and image rotation prediction~\cite{gidaris2018unsupervised}. \cite{wu2018unsupervised} propose a novel instance discrimination objective. The recent trend of contrastive learning~\cite{moco,simclr,ye2019unsupervised,byol,henaff2019data,tian2019contrastive} integrates data augmentations and instance discrimination by maintaining relative consistency between representations of an image and its augmented view. Clustering can also provide an effective addition~\cite{swav}. 
Recently, ~\cite{chen2021empirical} conduct contrastive learning using ViT~\cite{dosovitskiy2021an} and achieve impressive results.
As for the video domain, it is natural to exploit the temporal signals as the pretext task. Examples include predicting the future frame~\cite{srivastava2015unsupervised}, motion and appearance statistics~\cite{wang2019self}, speed~\cite{benaim2020speednet, wang2020self} and encodings~\cite{lotter2016deep,han2019video,han2020memory}, sorting frames or video clips~\cite{lee2017unsupervised,xu2019self,kim2019self,fernando2017self}. Recently, ~\cite{qian2020spatiotemporal} apply contrastive learning to videos with a temporal sampling strategy and temporally consistent spatial augmentation. 

\paragraph{Multimodal video.} Video is a natural source of multimodal data.
Multimodal self-supervised learning can be achieved by predicting whether a video has correspondence with an audio stream~\cite{arandjelovic2017look,arandjelovic2018objects,morgado2020audio,korbar2018cooperative}, cross-modality clustering~\cite{alwassel2019self}, and evolving losses~\cite{piergiovanni2020evolving}.
Recently, ~\cite{mmv} use contrastive loss to learn from video, audio and text; ~\cite{recasens2021broaden} learn to predict a broad view that spans a longer temporal context from a narrow view. 
\ours serves as a first work combining the strength of convolution-free Transformer and multimodal contrastive learning.

\section{Approach}
In this section, we introduce our convolution-free \ours architecture and elaborate on the self-supervised multimodal objectives for training \ours from scratch.

Figure~\ref{fig:overview} is an overview of the architecture. We feed each modality to a tokenization layer, where the raw input is projected to an embedding vector followed by a Transformer. There are two major settings: 1) The backbone Transformers are separate and have specific weights for each modality, and 2) The Transformers share weights, namely, there is a single backbone Transformer applied to any of the modalities. In either setting, the backbone extracts modality-specific representations, which are then mapped to common spaces to be compared with each other by contrastive losses. We describe each module in the following.

\subsection{Tokenization and Positional Encoding}
\ours operates on raw signals. The vision-modality input consists of 3-channel RGB pixels of video frames, the audio input is in the form of air density amplitudes (waveforms), and the text input is a sequence of words. We first define a modality-specific tokenization layer that takes as input the raw signals and returns a sequence of vectors to be fed to the Transformers. Besides, each modality has its own positional encoding, which injects the order of tokens into Transformers~\cite{vaswani2017attention}.
We partition an entire video clip of size $T\times H\times W$ to a sequence of $\ceil{T/t} \cdot \ceil{H/h} \cdot \ceil{W/w}$ patches, where each patch contains $t \times h \times w \times 3$ voxels. We apply a linear projection on the entire voxels in each patch to get a $d$-dimensional vector representation. This projection is performed by a learnable weight $\mW_{vp} \in \sR^{t\cdot h\cdot w\cdot 3\times d}$. This can be seen as a 3D extension of the patching mechanism proposed in~\cite{dosovitskiy2021an}. To encode the position of these patches, we define a dimension-specific sequence of learnable embeddings as follows:
\begin{equation}
\begin{split}
    \ve_{i,j,k} = {\ve_{\text{Temporal}}}_i +& {\ve_{\text{Horizontal}}}_j + {\ve_{\text{Vertical}}}_k,\\
    \mE_{\text{Temporal}} \in \sR^{\ceil{T/t}\times d},~~~\mE_{\text{Horizontal}} &\in \sR^{\ceil{H/h}\times d},~~~\mE_{\text{Vertical}} \in \sR^{\ceil{W/w}\times d}
\end{split}
\end{equation}
where $\ve_i$ is the $i$-th row of $\mE$. This scheme allows us to use $\ceil{T/t}+\ceil{H/h}+\ceil{W/w}$ positional embeddings to encode all the $\ceil{T/t}\cdot \ceil{H/h}\cdot \ceil{W/w}$ patches in a video clip.
The raw audio waveform is a 1D input with length $T'$, and we partition it to $\ceil{T'/t'}$ segments each containing $t'$ waveform amplitudes. Similar to video, we apply a linear projection with a learnable weight $\mW_{ap} \in \sR^{t'\times d}$ to all elements in a patch to get a $d$-dimensional vector representation. We use $\ceil{T'/t'}$ learnable embeddings to encode the position of each waveform segment.
For text, we first construct a vocabulary of size $v$ out of all words in our training dataset. For an input text sequence, we then map each word to a $v$-dimensional one-hot vector followed by a linear projection with a learnable weight $\mW_{tp} \in \sR^{v\times d}$. This is equivalent to an embedding dictionary lookup, which has been widely used in natural language understanding~\cite{mikolov2013efficient}.

\subsubsection{DropToken}
We introduce DropToken, a simple and yet effective strategy to reduce the computational complexity during training. Once we get the token sequence for the video or audio modality, we randomly sample a portion of the tokens and then feed the sampled sequence, not the complete set of tokens, to the Transformer. This is crucial for reducing the computational cost because a Transformer's computation complexity is quadratic, $O(N^2)$, where $N$ is number of tokens in the input sequence. Any effort on reducing the input length would reduce the number of FLOPs quadratically. This has an immediate impact on the wall clock time for training these models and makes it possible to host large models in limited hardware. We argue that instead of reducing the resolution or dimension of the raw inputs, it is better to keep a high-fidelity input and randomly sample the tokens via DropToken. DropToken is appealing especially with the raw video and audio inputs, which may contain high redundancies.

\subsection{The Transformer Architecture}
\label{tx-arch}
For simplicity, we adopt the most established Transformer architecture~\cite{devlin2018bert}, which has been widely used in NLP. Similar to ViT~\cite{dosovitskiy2021an}, we do not tweak the architecture so that our weights can be easily transferred to any standard Transformer implementation. We will briefly elaborate on the pipeline (also illustrated in Figure~\ref{fig:overview} middle panel) and refer the reader to~\cite{dosovitskiy2021an,devlin2018bert} for more details of the standard Transformer architecture. The sequence of input tokens to the Transformer follows the below formulation:
\begin{equation}
    \vz_{\text{in}} = [\vx_{\text{AGG}};~\vx_0\mW_{P};~\vx_1\mW_{P};\dots;~\vx_N\mW_{P}] + \ve_{\text{POS}}
\end{equation}
where $\vx_n$ is the input patches sequence and $\vx_{\text{AGG}}$ is the learnable embedding of a special aggregation token whose corresponding output in the Transformer ($z^0_{\text{out}}$) is used as the aggregated representation for the entire input sequence. This will be later used for classification and common space mapping. We use a standard self-attention~\cite{vaswani2017attention} as the Multi-Head-Attention (MHA) module, and GeLU~\cite{hendrycks2016gaussian} as the activation in the MLP layer. We also use Layer Normalization~\cite{ba2016layer} before the MHA and MLP modules. In our text model, we remove the position encoding $\ve_{\text{POS}}$ and add a learnable relative bias to each attention score of the first layer in the MHA module. This simple change makes our text model's weights directly transferable to the state-of-the-art text model T5~\cite{raffel2020exploring}.

\subsection{Common Space Projection}
We use common space projection and contrastive learning in that common space to train our networks. More specifically, given a video-audio-text triplet, we define a semantically hierarchical common space mapping that enables us to directly compare video-audio pairs as well as video-text pairs by the cosine similarity. As argued in~\cite{mmv}, such comparison is more feasible if we assume there are different levels of semantic granularity for these modalities. To achieve this, we define multi-level projections as follows:
\begin{equation}
\label{eq:common_space}
    \begin{split}
        \vz_{v,va} &= g_{v\rightarrow va}(\vz_{\text{out}}^{\text{video}}),~~~~~\vz_{a,va} = g_{a\rightarrow va}(\vz_{\text{out}}^{\text{audio}}) \\
        \vz_{t,vt} &= g_{t\rightarrow vt}(\vz_{\text{out}}^{\text{text}}),~~~~~~~~\vz_{v,vt} = g_{v\rightarrow vt}(\vz_{v,va})
    \end{split}
\end{equation}
where $g_{v\rightarrow va}$ and $g_{a\rightarrow va}$ are the projection heads to respectively map the video and audio Transformers' outputs to the video-audio common space $\gS_{va}$. Moreover, $g_{t\rightarrow vt}$ and $g_{v\rightarrow vt}$ project the text Transformer's outputs and the video embedding in the $\gS_{va}$ space to video-text common space, $\gS_{vt}$. This multi-level common space projection is depicted in Figure~\ref{fig:overview} (the rightmost panel). The main intuition behind this hierarchy is that different modalities have different levels of semantic granularity, so we should impose this as an inductive bias in the common space projection. Similar to~\cite{mmv}, we use a linear projection for $g_{a\rightarrow va}(.)$, $g_{t\rightarrow vt}(.)$, and $g_{v\rightarrow vt}(.)$, and a two-layer projection with ReLU in between for $g_{v\rightarrow va}(.)$. To ease the training, a batch normalization is used after each linear layer.

\subsection{Multimodal Contrastive Learning}
Inspired by~\cite{mmv,arandjelovic2017look,miech2020end}, we use Noise Contrastive Estimation (NCE) to align video-audio pairs and Multiple Instance Learning NCE (MIL-NCE) to align video-text pairs. The pairs are composed from different temporal locations in the video-audio-text stream. Positive pairs from two modalities are constructed by sampling their corresponding streams from the same location in the video, and negative pairs are constructed by sampling from any non-matching locations in the video~\cite{mmv}.
Concretely, given the common space specified in Section~\ref{eq:common_space}, the loss objectives can be written as follows:
\begin{equation}
        \text{NCE}(\vz_{v,va}, \vz_{a,va}) =
        -\log\left(\frac{\exp(\vz_{v,va}^{\top}\vz_{a,va}/\tau)}{\exp(\vz_{v,va}^{\top}\vz_{a,va}/\tau) + \sum_{z'\in\gN}{\exp({\vz'}_{v,va}^{\top}{\vz'}_{a,va}/\tau)}}\right), \label{eq:NCE}
\end{equation}

\begin{equation}
        \text{MIL-NCE}(\vz_{v,vt}, \{\vz_{t,vt}\}) =
        -\log\left(\frac{ \sum_{\vz_{t,vt}\in \gP}{\exp({\vz}_{v,vt}^{\top}{\vz}_{t,vt}/\tau)}}{\sum_{\vz_{t,vt}\in \gP}{\exp({\vz}_{v,vt}^{\top}{\vz_{t,vt}}/\tau)}+\sum_{z'\in\gN}{\exp({\vz'}_{v,vt}^{\top}{\vz'}_{t,vt}/\tau)}}\right),  \label{eq:mil-nce}
\end{equation}
where $\gN$ contains all non-matching pairs in a batch. In Equation~\ref{eq:mil-nce}, $\gP$ contains five text clips that are nearest neighbors to the video clip in time. $\tau$ is a temperature to adjust the softness of the objectives in distinguishing the positive pairs from the negative pairs.

The overall per-sample objective for training the entire \ours model end-to-end is as follows:
\begin{equation}
\label{eq:overall_loss}
    \gL = \text{NCE}(\vz_{v,va}, \vz_{a,va}) + \lambda\text{MIL-NCE}(\vz_{v,vt}, \{\vz_{t,vt}\}),
\end{equation}
where $\lambda$ balances the two losses. The model is optimized based on the back-propagation of the average loss calculated over a batch of samples.

\section{Experiments}
In this section, we first briefly describe the experimental setup for the pre-training and downstream evaluation, and then present the results and analytic interpretation of \ours in different tasks. We refer the reader to the Appendix for a more detailed description of all experimental settings.

\subsection{Experimental Setup}
\label{sec:exp_setup}
\paragraph{Pre-train:} we use a combination of AudioSet~\cite{audioset} and HowTo100M~\cite{howto100m} datasets to pre-train \ours --- we use only a subset of the HowTo100M dataset in compliance with Youtube's policies. Following~\cite{mmv}, we use video-audio-text triplets from HowTo100M clips while only using video-audio pairs from AudioSet. We sample 32 frames at 10 fps with a spatial size of $224\times224$ following a random crop, horizontal flip and color augmentation (details in \ref{sec:exp_setup_inputs}). Accordingly, we sample audio waveforms in sync at 48kHz. Both video and audio are normalized between [-1,1]. We use patch sizes of $4\times16\times16$ and $128$ for video and raw waveform tokenization, respectively (ablation in \ref{sec:ablation_inputs}). We use one-hot vectors to encode text sequences (capped to 16 tokens) with the vocabulary size of $2^{16}$. In all pre-training experiments, we use DropToken with drop rate 50\%. We train our models using the Adam optimizer~\cite{kingma2014adam} with a quarter-period cosine scheduled learning rate from 1$e$-4 to 5$e$-5 and 10k warmup steps. Optimization is performed on totally 500k steps with batch size 2048 (512 in exploration experiments). Following the previously established practice~\cite{mmv} for the projection to the common spaces $\gS_{va}$ and $\gS_{vt}$, we use $d_{va}=512$ and $d_{vt}=256$. We also use the temperature of $\tau=0.07$ and the weight of $\lambda=1$ in the loss in Equation~\ref{eq:overall_loss}. We use 4 network sizes in our experiments (details in \ref{sec:exp_setup_networks}). We use the Medium model (155M parameters) for our modality-agnostic variant (\ours-MA), and 3 variants for the modality-specific video-audio-text backbones: Base-Base-Small (BBS; 197M), Medium-Base-Small (MBS; 264M), and Large-Base-Small (LBS; 415M). Pre-training an MBS VATT with batch size 2048 on 256 TPUs (v3) takes less than 3 days. Pre-training with batch size 512 takes less than 1 day.
\paragraph{Downstream:} we evaluate the pre-trained \ours models on 4 major downstream tasks using a total of 10 datasets. We use UCF101~\cite{soomro2012ucf101}, HMDB51~\cite{hmdb}, Kinetics-400~\cite{kinetics400}, Kinetics-600~\cite{kinetics600}, and Moments in Time~\cite{monfort2019moments} for video action recognition. We use ESC50~\cite{esc} and AudioSet~\cite{audioset} for audio event classification, and we evaluate the quality of our video-text common space representations by zero-shot text-to-video retrieval on YouCook2~\cite{youcook2} and MSR-VTT~\cite{xu2016msr-vtt}. Finally, we evaluate the transferability of the vision backbone by fine-tuning it on ImageNet classification~\cite{deng2009imagenet}. Since HMDB51, UCF101, and ESC50 are very small datasets compared to the size of our networks, we only use them to train a linear classifier on top of the frozen pre-trained backbones. In our exploration experiments, we report linear classification accuracy and zero-shot video retrieval metrics. We refer to the Appendix for a detailed description of the datasets and the experimental setup.

\subsection{Results}
\input{tables/video_action_recognition_all.tex}
\subsubsection{Fine-tuning for video action recognition}
We fine-tune \ours's vision Transformer on Kinetics-400, Kinetics-600, and Moments in Time, three of the arguably most established large-scale datasets for video action recognition. We use the final checkpoints of four pre-train settings for these experiments: three modality-specific variations (\textit{LBS, MBS, BBS}), and one modality-agnostic (\textit{Medium}). Table~\ref{table:video-classification-all} shows the results compared with the state-of-the-art video models. On all three datasets, we achieve higher accuracy than previous works including TimeSFormer~\cite{bertasius2021space}, a recent effort in fine-tuning the ViT checkpoints obtained by \textit{supervised} pre-training. In contrast, our pre-training does not rely on any labels curated by humans. To the best of our knowledge, \ours provides the first vision Transformer backbone that is pre-trained from scratch using self-supervision on multimodal videos and achieves state-of-the-art results on video action recognition. It is also worth mentioning that fine-tuning VATT on the most recent Kinetics-700 dataset results in a top-1 accuracy of $72.7\%$, which outperforms the state-of-the-art top-1 accuracy of $72.4\%$ in~\cite{kondratyuk2021movinets}.

To further quantify how much the multimodal self-supervised pre-training helps in achieving these numbers, we train a variant from scratch without any pre-training and observe the top-1 and top-5 accuracies of $26.4\%$ and $51.8\%$ on Kinetics-400, respectively. The low accuracies verify the efficacy of our pre-training strategy for \ours. Finally, we find that VATT-MA-Medium, the modality-agnostic backbone shared by the video, audio, and text modalities, is on par with the modality-specific VATT-Base when fine-tuned for the video action recognition. This result is encouraging as it indicates the potential of unifying three data modalities by a \textit{single} Transformer backbone.

\subsubsection{Fine-tuning for audio event classification}
We fine-tune \ours's audio Transformer on AudioSet, which benchmarks the task of multi-label audio event classification. We use the final checkpoints of two pre-train settings: one modality-specific (BBS), and one modality-agnostic (Medium). Table~\ref{table:audio-classification} shows the results compared to state-of-the-art models. Following common practice~\cite{gemmeke2017audio,kong2019weakly}, we report mean Average Precision (mAP), Area Under Curve (AUC), and d-prime (based on AUC)~\cite{gemmeke2017audio}. Our audio Transformer consistently outperforms the existing CNN-based models in all metrics. More interestingly, fine-tuning the modality-agnostic backbone (VATT-MA-Medium) is on par with fine-tuning the modality-specific one (VATT-Base). To the best of our knowledge, \ours is the first Transformer that outperforms CNN-based models in audio event recognition. \ours operates on raw waveforms and does not utilize any handcrafted features.

\subsubsection{Fine-tuning for image classification}
In this section, we show that our pipeline is capable of transferring the learned knowledge into another domain by performing the image classification task, even though the models are pre-trained in the multimodal video domain.
We fine-tune the vision Transformer in \ours-BBS on ImageNet without any modification to the backbone architecture. Instead, to satisfy the voxel-to-patch layer we replicate the input image 4 times and feed it to the network. The network sees the input as a single-frame video clip and performs spatial self-attention. Table~\ref{table:image-classification} shows the results for fine-tuning the vision Transformer end-to-end on ImageNet. We can see that our pre-training leads to a significant boost in the accuracy compared to training from scratch. We also observe that even though the self-supervised pre-training happens in the video domain, we still achieve competitive results to the \emph{supervised} pre-training using large-scale \emph{image} data~\cite{dosovitskiy2021an}.

\subsubsection{Zero-shot text-to-video retrieval}
We feed video-text pairs to VATT-MBS, and extract representations in the $\gS_{vt}$ space. We then calculate the similarity between each video-text pair from YouCook2 and MSR-VTT. Given a text query, we rank the videos based on their similarities to the text. We then measure the recall for the correct video in the top-10 videos. We also measure the median of the rank of the correct video. Table~\ref{table:retrieval} compares our video retrieval results to two baselines. In our experiments we observe that the zero-shot retrieval results are heavily affected by the batch size and number of epochs, confirming the observation made in \cite{mmv}. That said, our model still delivers comparable results to MMV~\cite{mmv} while being pre-trained with a half number of epochs and a half batch size of theirs. We also experiment with a larger batch size 8192 and longer pre-training for 6 epochs, arriving at exactly the same results as MIL-NCE~\cite{miech2020end} on YouCook2 and the R@10 of 29.2 and MedR of 42 on MSR-VTT. We also notice that, probably due to the noisy nature of text transcripts, a sophisticated language model like ours is underrated. As shown in \cite{mmv}, using a simple linear projection would still perform reasonably well. It is worth exploring other, higher-quality text sources in future work.

\input{tables/aud_img_cls_vid_ret.tex}

\subsubsection{Feature visualization}
\begin{figure*}[t]
\centering
   \includegraphics[width=\linewidth]{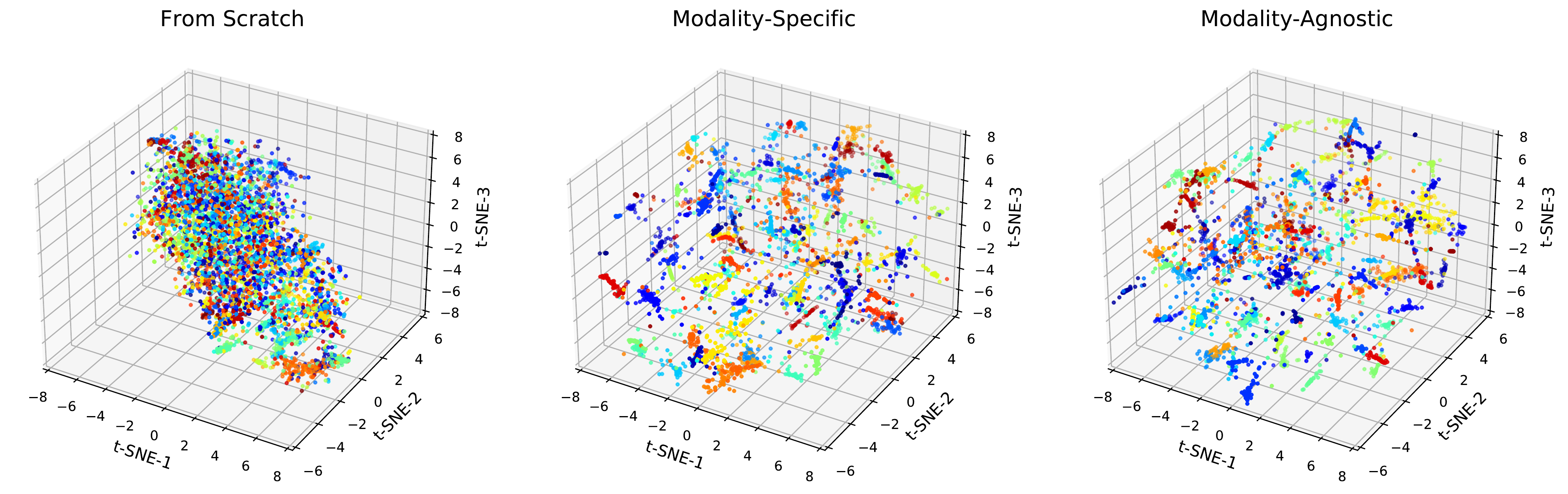}
   \caption{t-SNE visualization of the feature representations extracted by the vision Transformer in different training settings.  For better visualization, we show 100 random classes from Kinetics-400.}
\label{fig:tsne_k400}
\vspace{-2mm}
\end{figure*}

We take our modality-specific and modality-agnostic \ours fine-tuned on Kinetics-400 and visualize their output feature representations using t-SNE. For comparison, we also include the feature visualization of the vision Transformer trained from scratch on Kinetics-400. From Figure~\ref{fig:tsne_k400}, we observe that the fine-tuned \ours yields a much better separation than the model trained from scratch. Furthermore, it is worth noting that there is no clear difference between the modality-agnostic features and the modality-specific ones.

\begin{figure}[t]
    \centering
       \includegraphics[width=\columnwidth]{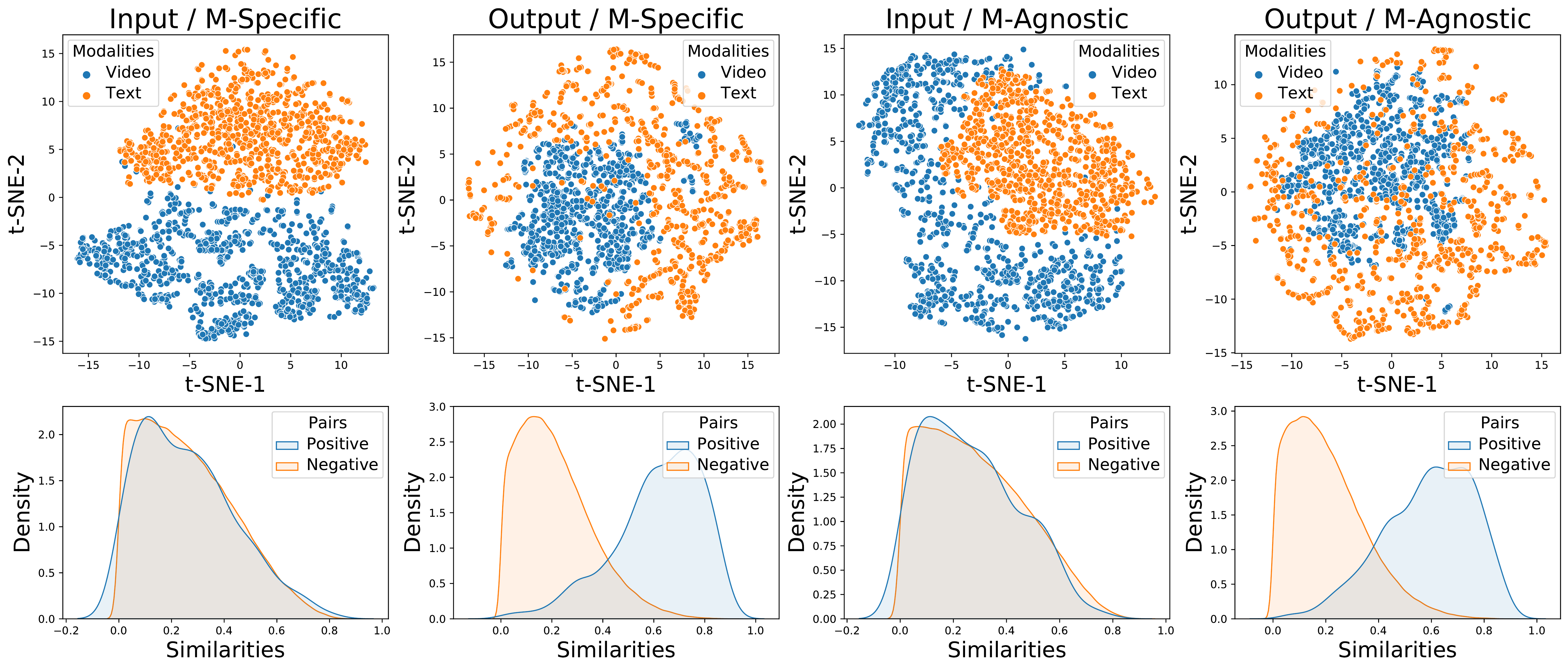}
       \caption{t-SNE visualization and distribution of pair-wise similarities of the input space vs.\ output space for modality-specific and modality-agnostic backbones when different modalities are fed.}
    \label{fig:analysis_inp_out}
    \vspace{-4mm}
\end{figure}

We further investigate the \ours backbones without any fine-tuning. We randomly choose 1k video clips from the YouCook2 dataset and store the representations from two points of a pre-trained \ours model. One is after the tokenization layer (input space of the Transformer), and the other is after the common space projection (output space), where the loss is computed. Figure~\ref{fig:analysis_inp_out}-top visualizes the representations, comparing modality-specific \ours to modality-agnostic \ours. Interestingly, we observe that the representations are slightly more mixed together in the modality-agnostic setting compared to the modality-specific ones, implying that the modality-agnostic backbone sees different modalities as different symbols describing the same concept. This is analogous to a unified language model in NLP that supports multiple languages.


To see how well \ours distinguishes positive video-text pairs from randomly sampled pairs, we calculate pair-wise similarities for all possible pairs and perform a Kernel Density Estimation (KDE) to visualize the distributions of the similarities of the positive pairs vs.\ negative pairs. We perform this procedure for both input and output spaces of the modality-specific and modality-agnostic backbones. Figure~\ref{fig:analysis_inp_out}-bottom shows the KDE curves of these similarities. We can see that \ours in both settings separates the positive and negative pairs in its output space. This verifies \ours's efficacy in learning a semantic common space for different modalities, even if we share the backbone across modalities.

\subsubsection{Model Activations}
\begin{figure}[t]
    \centering
       \includegraphics[width=0.7\linewidth]{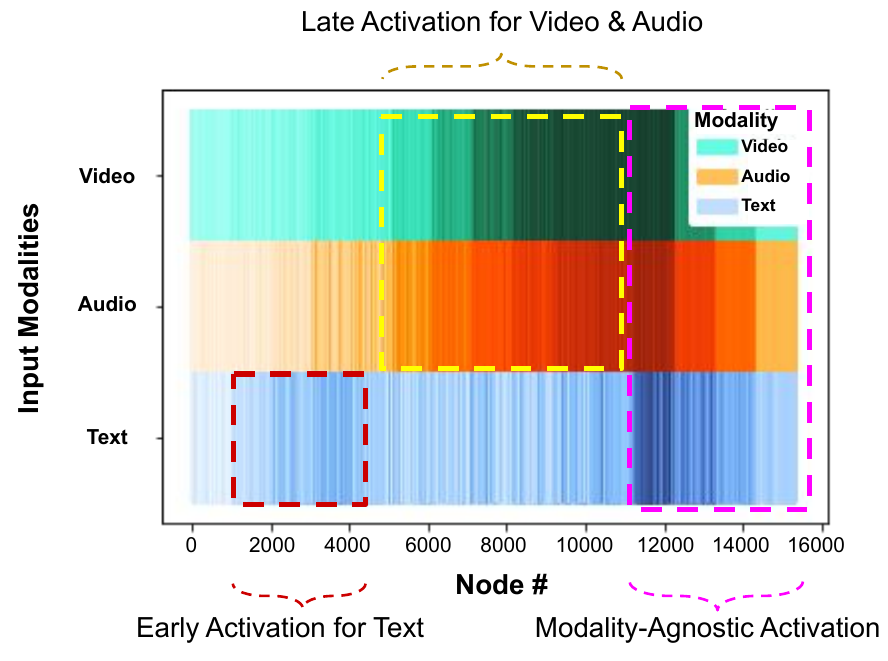}
       \caption{The average node activation across the Modality-Agnostic-Medium \ours while feeding a multimodal video-audio-text triplet to the model.}
    \label{fig:analysis_gelu_activation}
\end{figure}

\vspace{-2mm}
We measure the average activation of the modality-agnostic VATT when a full multimodal input is fed to the model. More specifically, we sample 100k short video clips from the test split of HowTo100M along with their corresponding audio and text and feed them to the model separately. For each modality, we calculate the average activation of each node at the output of the MLP module, before the residual addition (Figure \ref{fig:overview}-Transformer Encoder). Figure \ref{fig:analysis_gelu_activation} shows the average activations across all nodes in a Medium-size model. We observe that earlier nodes in the model are activated with the text inputs, while the middle-to-later nodes are activated with video and audio modalities. However, the nodes in the last layers of the network are activated with all modalities almost equally. This might suggest that the model allocates different nodes to certain modalities while reaching the same level of semantic perception for all modalities in the later layers. Such observation encourages further studies on the possibility of utilizing Mixture-of-Experts~\cite{shazeer2017outrageously,fedus2021switch,riquelme2021scaling} to increase the model's capacity for simultaneous multimodal perception. We leave this direction of research for future work.

\subsubsection{Effect of DropToken}
\vspace{-2mm}
We introduced a new method to reduce the redundancy in high-resolution data. To study the effect of the proposed DropToken method on downstream applications and the pre-training computation, we perform pre-training by randomly dropping $75\%$, $50\%$, $25\%$, and 0\% (no drop) of the tokens from the video and audio inputs. Table~\ref{table:drop-token-acc-pre-train} shows the accuracy of linear classification on HMDB51, UCF101, ESC50 and R@10 on YouCook2 and MSR-VTT vs.\ the drop rate along with GFLOPs during a forward call. We choose $50\%$ sampling rate for our large-scale pre-training as it offers a good trade-off between accuracy and computational costs. We then take the final checkpoint of the pre-trained \ours with $50\%$ DropToken rate and perform fine-tuning on Kinetics-400 at different DropToken rates and at different spatial and temporal resolutions to see how high-resolution inputs coupled with DropToken compare to low-resolution inputs with no tokens dropped during fine-tuning. Table~\ref{table:drop-token-acc-k400} shows the top-1 accuracy on Kinetics-400. We argue against using low-resolution inputs, which is the most common approach to reduce the computational cost during training. Instead, we suggest using high-resolution inputs with DropToken, whose accuracy and training cost are comparable to or better than low-resolution counterparts.
\input{tables/ablation_droptoken.tex}

\section{Conclusion and Discussion}
\label{sec:conclusion}
In this paper, we present a self-supervised multimodal representation learning framework based on Transformers. 
Our study suggests that Transformers are effective for learning semantic video/audio/text representations --- even if one model is shared across modalities --- and multimodal self-supervised pre-training is promising for reducing their dependency on large-scale labeled data. 
We show that DropToken can significantly reduce the pre-training complexity with video and audio modalities and have minor impact on the models' generalization. 
We report new records of results on video action recognition and audio event classification and competitive performance on image classification and video retrieval. 
Having these results, we still see some limitations in our work. Firstly, not all videos have organic audio or speech, while our approach depends on meaningful multimodal correspondences. Besides, the text modality currently consists of speech transcripts, which are noisy and sometimes sparse. Potential negative Societal Impacts are mainly concerned with applications. The models could be biased if one applies our approach to the multimodal videos that are not representative enough. Finally, our method is still demanding in computation, though we managed to avoid the need for human labels. Future work can improve upon these limitations.

\begin{ack}
We would like to thank Min-Hsuan Tsai, Jean-Baptise Alayrac, Andrew Audibert, Yeqing Li, Vidush Mukund, and the TensorFlow team for their help with codes, infrastructure, and insightful discussions.
\end{ack}

\setcitestyle{numbers}
\bibliographystyle{plainnat}
\bibliography{egbib}


\newpage
\input{appendix.tex}

\end{document}

%% file: math_commands.tex
\usepackage{amsmath,amsfonts,bm}

\def\eqref#1{equation~\ref{#1}}

\def\ceil#1{\lceil #1 \rceil}

\def\1{\bm{1}}

\def\ve{{\bm{e}}}

\def\vx{{\bm{x}}}
\def\vy{{\bm{y}}}
\def\vz{{\bm{z}}}

\def\mC{{\bm{C}}}

\def\mE{{\bm{E}}}

\def\mU{{\bm{U}}}
\def\mV{{\bm{V}}}
\def\mW{{\bm{W}}}

\DeclareMathAlphabet{\mathsfit}{\encodingdefault}{\sfdefault}{m}{sl}
\SetMathAlphabet{\mathsfit}{bold}{\encodingdefault}{\sfdefault}{bx}{n}

\def\gL{{\mathcal{L}}}

\def\gN{{\mathcal{N}}}

\def\gP{{\mathcal{P}}}

\def\gS{{\mathcal{S}}}

\def\sR{{\mathbb{R}}}



%% file: tables/video_action_recognition_all.tex
\begin{table}[t]
    \small
    \centering
    \begin{tabular}{@{}lcc|cc|cc|cc@{}}
    \toprule
    & \multicolumn{2}{c|}{\underline{Kinetics-400}} & \multicolumn{2}{c|}{\underline{Kinetics-600}} & \multicolumn{2}{c|}{\underline{Moments in Time}} \\
    \sc Method & \sc Top-1 & \sc Top-5 & \sc Top-1 & \sc Top-5 & \sc Top-1 & \sc Top-5 & \sc TFLOPs\\
    \midrule
    I3D~\cite{carreira2017quo} & 71.1 & 89.3 & 71.9 & 90.1 & 29.5 & 56.1 & -\\
    R(2+1)D~\cite{du2017r2+1d}  & 72.0 & 90.0 & - & - & - & - & 17.5\\
    bLVNet~\cite{fan2019blvnet} & 73.5 & 91.2 & - & - & 31.4 & 59.3 & 0.84\\
    S3D-G~\cite{xie2018rethinking} & 74.7 & 93.4 & - & - & - & - & -\\
    Oct-I3D+NL~\cite{chen2019drop} & 75.7 & - & 76.0 & - & - & - & 0.84\\
    D3D~\cite{stroud2020d3d} & 75.9 & - & 77.9 & - & - & - & -\\
    I3D+NL~\cite{wang2018nonlocal} & 77.7 & 93.3 & - & - & - & - & 10.8\\
    ip-CSN-152~\cite{tran2019video} & 77.8 & 92.8 & -  & - & - & - & 3.3\\
    AttentionNAS~\cite{wang2020attentionnas} & - & - & 79.8 & 94.4 & 32.5 & 60.3 & 1.0 \\
    AssembleNet-101~\cite{ryoo2019assemblenet} & - & - & - & - & 34.3 & 62.7 & -\\
    MoViNet-A5~\cite{kondratyuk2021movinets} & 78.2 & - & 82.7 & - & 39.1 & - & 0.29 \\
    LGD-3D-101~\cite{lgdnet} & 79.4 & 94.4 & 81.5 & 95.6 & - & - & -\\
    SlowFast-R101-NL~\cite{slowfast} & 79.8 & 93.9 & 81.8 & 95.1 & - & - & 7.0\\
    X3D-XL~\cite{feichtenhofer2020x3d} & 79.1 & 93.9 & 81.9 & 95.5 & - & - & 1.5\\
    X3D-XXL~\cite{feichtenhofer2020x3d} & 80.4 & 94.6 & - & - & - & - & 5.8\\
    TimeSFormer-L~\cite{bertasius2021space} & 80.7 & 94.7 & 82.2 & 95.6 & - & - & 7.14\\
    \midrule
    \ours-Base & 79.6 & 94.9 & 80.5 & 95.5 & 38.7 & 67.5 & 9.09\\
    \ours-Medium & 81.1 & \textbf{95.6} & 82.4 & 96.1 & 39.5 & \textbf{68.2} & 15.02\\
    \ours-Large & \textbf{82.1} & 95.5 & \textbf{83.6} & \textbf{96.6} & \textbf{41.1} & 67.7 & 29.80\\
    \midrule
    \ours-MA-Medium & 79.9 & 94.9 & 80.8 & 95.5 & 37.8 & 65.9 & 15.02\\
    \bottomrule
    \end{tabular}
    \vspace{2mm}
    \caption{Video action recognition accuracy on Kinetics-400, Kinetics-600, and Moments in Time.}
    \label{table:video-classification-all}
    \vspace{-6mm}
\end{table}

%% file: tables/aud_img_cls_vid_ret.tex
\begin{table}[t]
    \begin{minipage}[h]{0.4\textwidth}
        \small
        \centering
        \setlength{\tabcolsep}{2pt}
        \renewcommand{\arraystretch}{1.1}
        \begin{tabular}{@{}lcccc@{}}
        \toprule
            \sc Method & mAP & AUC & d-prime \\
        \midrule
        DaiNet~\cite{dai2017very} & 29.5 & 95.8 & 2.437\\
        LeeNet11~\cite{lee2017sample} & 26.6 & 95.3 & 2.371\\
        LeeNet24~\cite{lee2017sample} & 33.6 & 96.3 & 2.525\\
        Res1dNet31~\cite{kong2020panns} & 36.5 & 95.8 & 2.444\\
        Res1dNet51~\cite{kong2020panns} & 35.5 & 94.8 & 2.295\\
        Wavegram-CNN~\cite{kong2020panns} & 38.9 & 96.8 & 2.612\\
        \midrule
        \ours-Base & \textbf{39.4} & \textbf{97.1}  & \textbf{2.895}\\
        \midrule
        \ours-MA-Medium & 39.3 & 97.0  & 2.884\\
        \bottomrule
        \end{tabular}
        \vspace{2mm}
        \caption{Finetuning results for AudioSet event classification.}
        \label{table:audio-classification}
    \end{minipage}\hfill{}%
    \begin{minipage}[h]{0.55\textwidth}
        \footnotesize
        \centering
        \setlength{\tabcolsep}{8pt}
        \begin{tabular}{@{}lccc@{}}
        \toprule
            \sc Method & \sc Pre-training data & \sc Top-1 & \sc Top-5 \\
            
        \midrule
        iGPT-L~\cite{chen2020generative} & ImageNet & 72.6 & - \\
        ViT-Base~\cite{dosovitskiy2021an} & JFT & \textbf{79.9} & - \\
        \midrule
        \ours-Base & - & 64.7 & 83.9 \\
        \ours-Base & HowTo100M & 78.7 & 93.9\\
        \bottomrule
        \end{tabular}
        \vspace{1mm}
        \caption{Finetuning results for  ImageNet classification.}
        \label{table:image-classification}
        
        \footnotesize
        \centering
        \setlength{\tabcolsep}{1pt}
        \begin{tabular}{@{}lccccccc@{}}
        \toprule
            & & & \multicolumn{2}{c}{YouCook2} &  \multicolumn{2}{c}{MSR-VTT}\\ \cline{4-5} \cline{6-7}
            \sc Method & \sc Batch & \sc Epoch & R@10 & MedR & R@10 & MedR \\
            \midrule
        MIL-NCE ~\cite{miech2020end} & 8192 & 27 & \textbf{51.2} & \textbf{10} & \textbf{32.4} & \textbf{30} \\
        MMV ~\cite{mmv} & 4096 & 8 & 45.4 & 13 & 31.1 & 38 \\
        \midrule
        \ours-MBS & 2048 & 4 & 45.5 & 13 & 29.7 & 49 \\
        \midrule
        \ours-MA-Medium & 2048 & 4 & 40.6 & 17 & 23.6 & 67 \\
        \bottomrule
        \end{tabular}
        \vspace{1mm}
        \caption{Zero-shot text-to-video retrieval.}
        \label{table:retrieval} 
    \end{minipage}\hfill{}
    \vspace{-6mm}
\end{table}

%% file: tables/ablation_droptoken.tex
\begin{table}[h]
    \begin{minipage}[t]{0.45\textwidth}
        \footnotesize
        \centering
        \setlength{\tabcolsep}{2.5pt}
        \begin{tabular}{@{}lccccc@{}}
        \toprule
        & \multicolumn{4}{c}{DropToken Drop Rate} \\
        \cline{2-5}
         & $75\%$ & $50\%$ & $25\%$ & $0\%$ \\
        \midrule
        Multimodal GFLOPs & 188.1 & 375.4 & 574.2 & 784.8 \\
        \midrule
        HMDB51 & 62.5 & 64.8 & 65.6 & 66.4 \\
        UCF101 & 84.0 & 85.5 & 87.2 & 87.6 \\
        ESC50  & 78.9 & 84.1 & 84.6 & 84.9 \\
        YouCookII & 17.9 & 20.7 & 24.2 & 23.1 \\
        MSR-VTT & 14.1 & 14.6 & 15.1 & 15.2 \\
        \bottomrule
        \end{tabular}
        \vspace{2mm}
        \caption{Top-1 accuracy of linear classification and R@10 of video retrieval vs.\ drop rate vs.\ inference GFLOPs in the VATT-MBS.}
        \label{table:drop-token-acc-pre-train}
        \vspace{-6mm}
    \end{minipage}\hfill{}%
    \begin{minipage}[t]{0.5\textwidth}
        \footnotesize
        \centering
        \setlength{\tabcolsep}{3pt}
        \begin{tabular}{@{}l|ccccc@{}}
        \toprule
        Resolution/ & \multicolumn{4}{c}{DropToken Drop Rate} \\
        \cline{2-5}
        FLOPs & $75\%$ & $50\%$ & $25\%$ & $0\%$ \\
        \midrule
        $32\times224\times224$ & - & - & - & 79.9 \\
        Inference (GFLOPs) & - & - & - & 548.1 \\
        \midrule
        $64\times224\times224$ & - & - & - & 80.8 \\
        Inference (GFLOPs) & - & - & - & 1222.1 \\
        \midrule
        $32\times320\times320$ & 79.3 & 80.2 & 80.7 & 81.1 \\
        Inference (GFLOPs) & 279.8 & 572.5 & 898.9 & 1252.3 \\
        \bottomrule
        \end{tabular}
        \vspace{2mm}
        \caption{Top-1 accuracy of video action recognition on Kinetics400 using high-resolution inputs coupled with DropToken vs.\ low-resolution inputs.}
        \label{table:drop-token-acc-k400}
        \vspace{-6mm}
    \end{minipage}
\end{table}

%% file: appendix.tex
\appendix
\section{Appendix}

Appendix contains more detailed explanations about datasets (\ref{sec:datasets}) and the experimental setup (\ref{sec:exp_setup}) for both pre-training and downstream tasks. We also cover linear evaluation results compared to state-of-the-art (\ref{sec:linear_eval}) and an ablation study on the input parameters (\ref{sec:ablation_inputs}).

\subsection{Datasets}
\label{sec:datasets}
\subsubsection{Pre-training}
Following~\cite{mmv,miech2020end}, we use HowTo100M~\cite{howto100m} and AudioSet~\cite{audioset} to pre-train \ours. The former contains 1.2M unique videos, each providing multiple clips with audio and narration scripts resulting in 136M video-audio-text triplets in total. The narration scripts are extracted from speech audio using an off-the-shelf ASR. We use a subset of HowTo100M to comply with Youtube's policies, which results in having almost 1M unique videos and less than 100M clips. AudioSet consists of 10-second clips sampled from two million videos from YouTube. The dataset contains a variety of audio events with their corresponding video without any narration, so we do not have any text input from this dataset. We do not use any labels from the datasets. We uniformly sample clips from these datasets; a mini-batch in the pre-training contains samples from both datasets. In order to fill in the empty text in AudioSet, we feed a sequence of zeros to the text Transformer and exclude those samples from the MIL-NCE loss. 

\subsubsection{Downstream}
We evaluate the pre-trained \ours on a set of diverse, representative downstream tasks to test different aspects of the learned representations. 

\paragraph{Video action recognition:} We evaluate the visual representations on UCF101~\cite{soomro2012ucf101} (101 classes, 13,320 videos), HMDB51~\cite{hmdb} (51 classes, 6,766 videos), Kinetics-400~\cite{kinetics400} (400 classes, 234,584 videos), Kinetics-600~\cite{kinetics600} (600 classes, 366,016 videos), and Moments in Time~\cite{monfort2019moments} (339 classes, 791,297 videos). Since UCF101 and HMDB51 are small datasets compared to the size of our model, we freeze the vision backbone and use its outputs to train a linear classifier. We use the split \#1 results of the two datasets as a reference in our design exploration.
For Kinetics-400, Kinetics-600, and Moments in Time, we fine-tune our vision backbone initialized from the pre-trained checkpoint.

\paragraph{Audio event classification:} We use ESC50~\cite{esc} (50 classes, 2000 audio clips) and AudioSet~\cite{audioset} (527 classes, $\sim$2M audio clips) to evaluate our audio Transformer on audio event classification. We use ESC50 to train a linear classifier on top of the frozen audio Transformer. We use the split \#1 results of this dataset as a reference in our design exploration. We also use AudioSet to fine-tune our audio backbone initialized from the pre-trained checkpoint.

\paragraph{Zero-shot video retrieval:} We evaluate the quality of our video-text common space representations by \textit{zero-shot} text-to-video retrieval on two of the most established datasets in this area: YouCook2~\cite{youcook2} and MSR-VTT~\cite{xu2016msr-vtt} with 3.1k and 1k video-text pairs, respectively. We follow the same evaluation pipeline described in~\cite{mmv} and report the Recall at 10 (R@10).

\paragraph{Image classification:} Although there exists a domain gap between images and the video datasets used for pre-training \ours, we test the learned vision Transformer in the image domain. We fine-tune the last checkpoint of the vision Transformer on ImageNet~\cite{deng2009imagenet} with no modification to our architecture or the tokenization pipeline. We will elaborate on this in the sequel.

\subsection{Experimental Setup}
\label{sec:exp_setup}
\subsubsection{Inputs}
\label{sec:exp_setup_inputs}
During pre-training, we sample 32 frames at 10 fps for both pre-training datasets. For these frames, we randomly crop a temporally consistent spatial region whose relative area is in the range of [0.08, 1] and its aspect ratio in [0.5, 2]. These crops are then resized to $224\times224$, followed by a horizontal flip and color augmentation. The color augmentation follows~\cite{mmv} and randomizes brightness (max delta = 32/255), saturation (max delta = 0.4),
contrast (max delta=0.4), and hue (max delta=0.2). We clip values to ensure the RGB is in [0, 1]. The audio waveforms are sampled in sync with the video frames at 48kHz. Both video and audio inputs are normalized between [-1, 1] for numerical stability. We use patch sizes of $4\times16\times16$ and $128$ for video and raw waveform tokenization, respectively. We use one-hot vectors to encode text sequences with the vocabulary size of $2^{16}$, which is the same as word2vec~\cite{mikolov2013efficient}. The resulting sequence retains a maximum of 16 words by either clipping or padding. We use DropToken with a drop rate of $50\%$ during pre-training. For video fine-tuning and evaluation, 32 frames with a temporal stride of 2 are sampled at 25 fps (2.56 seconds) with a crop size of $320\times320$ (with similar video augmentation during pre-training), and we do not drop any tokens. We do not change the input size for audio and text during evaluation.

\subsubsection{Network setup in \ours}
\label{sec:exp_setup_networks}
We use the same Transformer architecture described in the main paper with various sizes shown in Table~\ref{table:arch-size}. We use the Medium model for our modality-agnostic variant (\ours-MA). For the experiments with modality-specific Transformers, we use the Small and Base models for the text and audio modalities, respectively, while varying the model sizes for the video modality. This results in 3 variants for the modality-specific video-audio-text backbones: Base-Base-Small (BBS), Medium-Base-Small (MBS), and Large-Base-Small (LBS).
\input{tables/architecture_size.tex}

\subsubsection{Projection heads and contrastive losses} 
We use $d_{va}=512$ and $d_{vt}=256$ for the projection to the common spaces $\gS_{va}$ and $\gS_{vt}$, respectively. We normalize the vectors before calculating the NCE and MIL-NCE objectives and use the temperature of $\tau=0.07$ and the weight of $\lambda=1$ in the loss defined in the paper. We choose these values following the previously established practice~\cite{mmv}; we may achieve better results by varying these hyper-parameters.

\subsubsection{Pre-training setup} 
We pre-train \ours from scratch using Adam~\cite{kingma2014adam} with an initial learning rate of 1$e$-4, 10k warmup steps, 500k steps in total, a batch size of 2048, and a quarter-period cosine schedule to anneal the learning rate from 1$e$-4 to 5$e$-5. In the exploration experiments, we use a batch size of 512 while keeping the rest of the training parameters the same. Our pipeline is implemented in Tensorflow (v2.4), and our models are trained for 3 days using 256 TPUs (v3).

\subsubsection{Video fine-tuning setup} 
For video action recognition, we use the SGD with a momentum of 0.9 and an initial learning rate of $0.005$, 2.5k warmup steps, a batch size of 64, 100k steps in total, and a half-period cosine schedule to anneal the learning rate to 0. We use label smoothing with smoothing factor $\alpha = 0.1$. The video frame resolution is $320\times320$, which results in an increase in the number of positional encoding weights. This increase is due to the fact that, in the pre-train time, we have 8+14+14 positional encoding buckets, while 8+20+20 positional buckets are required to completely encode $320/16$ horizontal and $320/16$ vertical locations in fine-tune. To generate the new positional embeddings, we create a new set of positional encoding buckets by bi-cubic interpolation from the original buckets. After this step, we fine-tune the entire network, including the positional encoding buckets, end-to-end. We tried fixed positional embeddings (solely based on interpolation for the missing locations) and did not observe significant improvements. We uniformly sample 4 clips to cover the entire 10 seconds of the video and apply a standard 3-crop evaluation following~\cite{slowfast}. We average the logits across the resulting 12 views before having the final class predictions.

\subsection{Audio fine-tuning setup} 
For audio event classification, we use the SGD with a momentum of 0.9, an initial learning rate of $0.2$, 5k warmup steps, a batch size of 1024, 50k steps in total, and a half-period cosine schedule to anneal the learning rate to 0. We observe that increasing the effective receptive field improves the overall performance. We suggest that this might be due to the fact that the AudioSet annotations are multi-label and each event might occur in different temporal positions. Hence, we employ the duration of 6.4s with 24kHz sampling rate (153.6k total input samples). Similar to~\cite{kong2020panns}, we use mixup~\cite{zhang2017mixup} on input-label ($\vx$-$\vy$) pairs in a mini-batch as below:
\begin{equation*}
    \vx = \alpha\vx_1 + (1-\alpha)\vx_2,~~~~~~\vy = \alpha\vy_1 + (1-\alpha)\vy_2,
\end{equation*}
where the input-label pairs are randomly sampled from a mini-batch, and the mixing rate $\alpha$ is sampled from a Beta$(5,5)$ distribution. We also perform data balancing by penalizing the loss value of a sample with the inverse of the per-batch number of repetitive labels it carries. This is crucial for avoiding over-fitting since AudioSet has a long-tailed distribution, and a few dominant classes may disrupt the training~\cite{kong2020panns}.

\subsubsection{Image fine-tuning setup} 
We finetune the pre-trained \ours on ImageNet for 50 epochs with $384 \times 384$ input resolution, 512 batch size, SGD with momentum of 0.9, cosine learning rate decay with an initial learning rate of 8$e$-2, and label smoothing of 0.1. No weight decay is used.

\subsubsection{Linear evaluation setup} 
We use a linear classifier with fixed backbones across all datasets and tasks. We observe that using matrix factorization on the classifier weight~\cite{rendle2010factorization} leads to a more stable result across experiments. More specifically, we use a factorized weight $\mC=\mU\mV \in \sR^{d\times c}$, where $\mU \in \sR^{d\times n}$ and $\mV \in \sR^{n\times c}$ are learnable weights. During training this classifier, we randomly choose a subset of the $n$ components in $\mU$ and $\mV$, hence leading to a low-rank classifier weight, $\mC$. The classifier weight, $\mC$, is trained using the Adam optimizer with a learning rate of 5$e$-4, a batch size of 64, a total of 50k training steps, and a sampling rate of 10\% on its $n=128$ components.

\subsubsection{Zero-shot retrieval setup} 
For zero-shot text-to-video retrieval, we use the 1k split of MSR-VTT and the entire test split of YouCook2 as the pool for retrieval. We use $224\times224$ central crops for 32 frames with a temporal stride of 2 sampled at 25 fps. Since each input clip covers 2.56 seconds, and the full clip length is 10 seconds, we average the embeddings over 4 uniformly sampled clips before calculating the similarity with a text query's embedding. We $\ell_2$-normalize each vector to assure that a dot product results in the cosine similarity.

\subsection{Linear evaluation on frozen \ours}
\label{sec:linear_eval}
We also test \ours's ability to generalize to other datasets when the entire backbone is frozen. In this setting, we focus on the video and audio modalities and train a linear classifier on the outputs of the frozen backbones. In addition to the low-rank classifier (LRC) described in Section~\ref{sec:exp_setup}, we also report the results of a SVM classifier following the same pipeline as~\cite{mmv}. Table~\ref{table:linear-eval} shows the performance of our model on three datasets. We observe that \ours does not outperform the best CNN counterparts in~\cite{mmv}, and achieves comparable numbers to other baselines. This could suggest that \ours's backbones learn less-linearly-separable feature, especially given that the contrastive estimation head includes non-linear projections.

\input{tables/linear_eval}

\subsection{Ablation study on input parameters}
\label{sec:ablation_inputs}
Since \ours takes raw multimodal signals as inputs, the choice of input size and how they are patched has a significant impact on the final performance. First, we alter the frame crop size and the number of sampled frames from each video clip while keeping the patch size fixed to $5\times16\times16$.  Table~\ref{table:video_input} shows that using a small frame crop size and a larger number of frames hurts the video-related results, but it does not significantly change the audio classification numbers.

\input{tables/ablation_video_input.tex}

Then, we keep the best frame size ($32\times224\times224$) and vary the video patch size. We find going beyond $4\times16\times16$ along either the time or spatial dimensions is not helpful. We avoid patches that are smaller than $4\times16\times16$ because of the significantly increaseed wall clock time in experiments.

Finally, we compare different audio patch sizes and perform an experiment using spectrograms, as opposed to the raw waveforms, as audio input. The goal is to see how the raw waveforms compare to the handcrafted spectrograms. We use the MEL spectrogram with 80 bins, the STFT length of 42 ms, and the STFT step of 21 ms following a similar setup in~\cite{mmv}. Tables~\ref{table:audio_input} summarize the results, in which we observe that the patch size of 128 gives rise to the best waveform-based results, and using spectrogram does not lead to any conclusive improvement. The experiment with the spectrograms demonstrates that \ours is able to learn semantic representations from raw audios. To the best of our knowledge, this is the first time that raw audio waveforms are used for multimodal self-supervised learning.

\input{tables/ablation_audio_input.tex}


%% file: tables/architecture_size.tex
\begin{table}[h!]
    \small
    \centering
    \setlength{\tabcolsep}{2pt}
    \begin{tabular}{@{}lcccccc@{}}
    \toprule
    Model & Layers & Hidden Size & MLP Size & Heads & Params \\
    \midrule
    Small & 6 & 512 & 2048 & 8 & 20.9 M \\
    Base & 12 & 768 & 3072 & 12 & 87.9 M \\
    Medium & 12 & 1024 & 4096 & 16 & 155.0 M \\
    Large & 24 & 1024 & 4096 & 16 & 306.1 M \\
    \bottomrule
    \end{tabular}
    \vspace{2mm}
    \caption{Details of the Transformer architectures in \ours.}
    \label{table:arch-size}
\end{table}

%% file: tables/linear_eval.tex
\begin{table}[ht]
    \small
    \centering
    \begin{tabular}{@{}lccccc@{}}
    \toprule
        \sc Method & UCF101 & HMDB51 & ESC50\\
    \midrule
    MIL-NCE~\cite{miech2020end} & 83.4 & 54.8 & - \\
    AVTS~\cite{korbar2018cooperative} & - & - & 82.3 \\
    XDC~\cite{alwassel2019self} & - & - & 84.8 \\
    ELo~\cite{piergiovanni2020evolving} & - & 64.5 & - \\
    AVID~\cite{smith2019avid} & - & - & \textbf{89.2} \\
    GDT~\cite{patrick2020multi} & - & - & 88.5 \\
    MMV~\cite{mmv} & \textbf{91.8} & \textbf{67.1} & 88.9 \\
    \midrule
    \ours-Medium + SVM & 89.2 & 63.3 & 82.5\\
    \ours-Medium + LRC & 89.6 & 65.2 & 84.7\\
    \midrule
    \ours-MA-Medium + LRC & 84.4 & 63.1 & 81.2\\
    \bottomrule
    \end{tabular}
    \vspace{2mm}
    \caption{Linear evaluation results for video action recognition on UCF101 and HMDB51 and audio event classification on ESC50. MA refers to the Modality-Agnostic backbone.}
    \label{table:linear-eval}
    \vspace{-4mm}
\end{table}

%% file: tables/ablation_video_input.tex
\begin{table}[h!]
    \small
    \centering
    \begin{tabular}{@{}lcccccc@{}}
    \toprule
    Frame Size & Patch Size & UCF & HMDB & YC2 & MSRVTT & ESC \\
    \midrule
    32$\times$224$\times$224 & 4$\times$16$\times$16 & \textbf{87.8} & \textbf{67.7} & \textbf{27.53} & \textbf{17.99} & \textbf{87} \\
    \cdashlinelr{1-7}
    32$\times$200$\times$200 & 5$\times$16$\times$16 & 87.16 & 67.08 & 23.98 & 17.84 & 86.25 \\
    32$\times$224$\times$224 & 5$\times$16$\times$16 & 87.74 & 67.6 & 27.47 & 17.96 & \textbf{87} \\
    64$\times$224$\times$224 & 5$\times$16$\times$16 & 86.57 & 63.09 & 18.52 & 12.5 & 86.25 \\
    \cdashlinelr{1-7}
    32$\times$224$\times$224 & 8$\times$16$\times$16 & 86.52 & 65.64 & 23.43 & 16.14 & 84 \\
    32$\times$224$\times$224 & 8$\times$32$\times$32 & 82.68 & 60.73 & 15.27 & 13.79 & 87 \\
    \bottomrule
    \end{tabular}
    \vspace{2mm}
    \caption{Effect of video frame and patch size on downstream results.}
    \label{table:video_input}
    \vspace{-2mm}
\end{table}

%% file: tables/ablation_audio_input.tex
\begin{table}[h!]
    \small
    \centering
    \setlength{\tabcolsep}{3pt}
    \begin{tabular}{@{}lccccccc@{}}
    \toprule
    Input & Patch Size & UCF & HMDB & YC2 & MSRVTT & ESC \\
    \midrule
    Waveform & 128 & \textbf{88.14} & \textbf{68.13} & 25.72 & \textbf{17.31} & \textbf{87.75} \\
    Waveform & 256 & 87.74 & 66.1 & 24.19 & 16.55 & 83.75 \\
    Waveform & 512 & 87.21 & 67.34 & \textbf{26.11} & 16.91 & 82.5 \\
    Waveform & 1024 & 86.41 & 66.36 & 24.46 & 16.38 & 82.5 \\
    \midrule
    Spectrogram & 16 $\times$ 5 & \textbf{88.3} & 67.52 & \textbf{26.62} & 16.86 & \textbf{88} \\
    \bottomrule
    \end{tabular}
    \vspace{2mm}
    \caption{Effect of the audio input type and patch size on downstream results.}
    \label{table:audio_input}
    \vspace{-2mm}
\end{table}